# Data Augmentation and List Integration for Improving Domain-Specific Sinhala-English-Tamil Statistical Machine Translation

Aloka Fernando    Gihan Dias    Surangika Ranathunga
Department of Computer Science and Engineering
University of Moratuwa
Katubedda 10400, Sri Lanka
{alokaf, gihan, surangika}@cse.mrt.ac.lk

*Abstract*—Out of vocabulary (OOV) is a problem in the context of Machine Translation (MT) in low-resourced languages. When source and/or target languages are morphologically rich, it becomes even worse. Bilingual list integration is an approach to address the OOV problem. This allows more words to be translated than are in the training data. However, since bilingual lists contain words in the base form, it will not translate inflected forms for morphologically rich languages such as Sinhala and Tamil. This paper focuses on data augmentation techniques where bilingual lexicon terms are expanded based on case-markers with the objective of generating new words, to be used in Statistical machine Translation (SMT). This data augmentation technique for dictionary terms shows improved BLEU scores for Sinhala-English SMT.

## I. INTRODUCTION

Machine Translation (MT) [1] is a sub-field of AI and Computational Linguistics, enabling the conversion of text, from one natural language, referred to as a source language to another natural language, referred as the target language. The MT research dates back to 1940 [2] and has gained a drastic improvement in the recent years. This is mainly due to the availability of language supported tools and publicly available resources such as parallel corpus, lexicons etc.

Data-driven MT techniques [3] mainly depend on parallel corpora. Under the data-driven approach, a model is trained on a parallel corpus, which can perform the translation from the source language to the target language. The data-driven approach can be further classified as example-based MT, statistical Machine Translation (SMT) and Neural Machine Translation (NMT). SMT addresses the MT challenge by learning a statistical model, where input parameters are derived from the analysis of parallel bilingual text corpora. Neural machine translation [4] models fit a single model rather than a pipeline of fine-tuned models and currently achieve state-of-the-art results. While NMT performance is appealing, in the context of low-resourced languages, SMT techniques [5] have continued to produce comparable or better results for the same.

Sri Lanka being a multi-ethnic country has main languages as Sinhala, Tamil and English. Therefore formal government documents are made available in all three languages. Currently it is human translators who perform the translation task hence its time consuming and labour intensive. To better cater to this need, an MT system among Sinhala-Tamil-English languages is a timely requirement.

For MT, the parallel corpus is critical, and Sinhala-Tamil-English being low-resourced languages it is a bottleneck. In this context [6] the training data would have a low coverage of the vocabulary of the respective natural language. Hence the Translation Model (TM) would be restricted to the limited-vocabulary, giving rise to the OOV problem. Farhath et al [7] has improved the Sinhala-Tamil translation with list integration. The lists used in that research contained terms are in its base-singular form, but when the term occurs in its inflected form still the TM will not translate the inflected terms.

Consider the following example in Table I,

TABLE I. Translation Example

| A | si sent | ඉඩම් ප්‍රතිසංස්කරණ කොමිෂන් සභාව |
|---|---|---|
|   | en sent (translation) | Land Reform Commission |
| B | si sent | ඉඩම් ප්‍රතිසංස්කරණ කොමිෂන් සභාවෙන් |
|   | en sent (translation) | Land Reform Commission සභාවෙන් |

Although sentence (A) is correctly translated into English, when the last word is in its inflected form in sentence (B), ie. සභාවෙන් the TM fails for the last word.

Therefore for MT involving morphologically rich languages, such as Sinhala and Tamil, the approach for OOV handling should also serve the word inflections as well.

This paper presents a data augmentation technique to improve SMT of Sinhala Tamil and English similar to Farhath et al. [8]. However, in addition to using bilingual dictionaries and domain-specific glossaries in their base form, we inflected the terms appearing in these lists using language-specific rules, hence expanding the same. Results show that this list expansion helps in reducing the OOV rate in the SMT output, while inflection of single words results in an increase in BLUE for Sinhala-English SMT.

Although list integration increases the BLEU and addresses the OOV problem, the Reordering Model (RM) [9] would not be able to position it correctly in the translated output. This disturbs the fluency of the translation output. Therefore the lists were integrated by filtering the list terms which exist in the training corpus, thereby minimising the effect of the positioning (reordering) issues in the translation output. Results prove that such filtration gives rise to improvement in BLEU score.

The rest of the paper is organized as follows. Section II discusses the existing research work in the domain of MT for Sinhala-Tamil-English languages and data augmentation techniques towards SMT improvements, section III on the SMT experimental setup, parallel corpus used in the experiments and the data augmentation and filtration techniques towards improving the translation. Section IV is a discussion on the translation scores and observations made, and finally Section 7 concludes the paper by presenting conclusions and future work.

II. RELATED WORK

*A. Statistical Machine Translation (SMT)*

Statistical Machine Translation (SMT) obtains the best translation, $e_{best}$ by maximising the conditional probability of the foreign sentence $p(e \vee f)$ given the Phrase-Translation model $p_\phi(f \vee e)$, Language Model $p_{LM}(e)$, Distortion Model $p_D(e,f)$ and Word Penalty $\omega^{length}(e)$ [10]

$$e_{best} p(e \vee f) = argmax_e p(e \vee f) \quad (1)$$

$$argmax_e p(e \vee f) = argmax_e p_\phi(f \vee e) x p_{LM}(e) x p_D(e,f) x \omega^{length}(e) \quad (2)$$

State-of-the-art SMT systems rely on (a) large bilingual corpora to train the translation model $p(f|e)$ and (b) monolingual corpora to build the language model, $p_{LM}(e)$.

In phrase-based models, longer translation units (more than a single word) are considered to be the atomic units and the method used to identify the mapping between target phrase to the source phrase is termed symmetrizing. To create symmetrization, word alignment is trained in either direction (source-to-target and target-to-source) separately.

The Language Model (LM) models the fluency of the proposed target sentence with greater probabilities appointed to sentences that are more common in the natural language. An n-gram LM would be used in the model to score the likely-hood of the translation hypothesis.

*B. Sinhala-Tamil-English Machine Translation*

There had been few attempts for MT among Sinhala-Tamil-English language pairs. Although English language is rich in terms of linguistic resources, for Sinhala and Tamil language-resources such as tokenizers, lemmatizers, morphological analysers, and POS taggers are still in their early stages, and a dependency parser is still not available for Sinhala [11]. Therefore the linguistic-based processing which can be incorporated into the translation models are challenging.

Early attempts of open-domain Sinhala-Tamil translation have been by Weerasinghe [12] while Farhath et al.[8] addressed the translation task as a closed-domain Sinhala-Tamil problem for domain official government documents. Further, compared to SMT, NMT research has not performed well in the MT task for Sinhala, Tamil languages [13], [14] mainly owing to a low-parallel data.

When it comes to Sinhala-English translation, there has been initial work but these have not produced satisfactory results. Further the attempts have been rule-based, supported by morphological information [15], and using SMT [16].

Tamil-English MT approaches have been mainly on Phrase-based SMT [17] Afterwards MT improvements had been in the area of Factored SMT considering the morphological information [18]. Further work involves implementing suffix-separation rules for Tamil-English languages as a preprocessing technique, incorporated into the phrase-based as well as hierarchical SMT [19] The Factored-based model had shown promising results for Tamil-English language pairs.

*C. List Integration*

List integration techniques have been incorporated into the translation process as static and dynamic techniques depending on where in the translation pipeline it influences [20].

Static integration would incorporate the list of items at different steps of TM creation, such as parallel corpus creation, phrase extraction, phrase scoring, and phrase table generation, or at system tuning when adjusting weights.

Bouamor et al. [20] conclude that adding lists to the parallel corpus statically improves the translation quality for English-French pair.

In dynamic integration, the list affects the translation process (decoding). This does not alter the pre-defined models or weights, however it is handled as a pre or post processing, without the translation process [21].

Carpuat and Diab [22] evaluated static (terminologies treated as single units), as well as dynamic (special feature used to indicate the presence of terminology) integrations. Improvements in the quality of translation were noticed in both, while static integration outperformed dynamic integration.

Therefore the lists would be integrated as static, towards improvement of the translation.

*D. Data Augmentation*

Data augmentation [9] is a widely adopted technique towards improving MT. Data augmentation approaches the OOV problem by introducing more vocabulary into the TM.

Recent work on data augmentation by Fadaee et al [23] considers augmenting low-frequency words in the training corpus by providing novel conexes in a synthetically generated sentences.

Following the above work, Tennage et al [13] has incorporated syntactic linguistic features in producing the synthetic sentences in MT. Parts of Speech (POS) tagging and morphological analysis have been used as syntactic features to prune the generated synthetic sentences that only comply with the language syntax.

Although these work has contributed to the improvement in the translation output the coverage of OOV is only limited to rare words.

The main problem specific for Sinhala and Tamil languages being the languages are rich in terms of morphology. Therefore the words are highly inflected. Therefore focusing on a technique which will augment more inflected terms would be effective towards addressing the OOV problem.

### III. METHODOLOGY

#### A. Sinhala-Tamil-English Parallel Corpus

We use a trilingual parallel dataset specific to government documents. Statistics are as per II. These have been cleaned and aligned manually. Hence the parallel corpus is a quality dataset.

Randomly sentences were selected for testing, validation and training sets. Table III outlines the Training-Validation-Testing split in-terms of the number of sentences, words and the unique words. This trilingual parallel corpus is used for the *Sinhala-English* and *Tamil-English* SMT experiments.

For the *Sinhala-Tamil* translation, as the training set, a combined dataset has been used. Specifically, the 13K Sinhala-Tamil sentences from the trilingual parallel corpus and the Sinhala-Tamil parallel corpus used by Farhath et al [7]. The corpus statistics are as per table IV.

TABLE II. PARALLEL CORPUS STATISTICS

| Document Category | No of Sentences |
|---|---|
| Annual Reports | 6,196 |
| Procurement | 928 |
| Websites | 6,411 |
| Committee Reports | 2,481 |
| Government Act | 401 |

TABLE III. TRAINING-VALIDATION-TESTING SET WISE STATISTICS FOR SINHALA-ENGLISH AND TAMIL-ENGLISH TRANSLATION

| Data Set | No. Sent | | Language | | |
|---|---|---|---|---|---|
| | | | En | Si | Ta |
| Train Set | 13191 | Total Words | 103876 | 100446 | 84341 |
| | | Unique Words | 11076 | 12325 | 19471 |
| Tuning Set | 1623 | Total Words | 23578 | 22721 | 18472 |
| | | Unique Words | 4788 | 5351 | 7645 |
| Testing Set | 1603 | Total Words | 19248 | 18513 | 15502 |
| | | Unique Words | 4237 | 4520 | 6171 |

TABLE IV. TRAINING SET WISE STATISTICS FOR SINHALA-TAMIL TRANSLATION

| Data Set | Sent. | No of Words | | No of Unique Words | |
|---|---|---|---|---|---|
| | | Si | Ta | Si | Ta |
| Training Set | 40,895 | 432128 | 393735 | 25661 | 47977 |

The monolingual corpus has been used in the Language Model. In the SMT pipeline the Language Model would determine the fluency of the translation output.

Table V outlines the monolingual corpus statistics used in the experiments.

TABLE V. MONOLINGUAL CORPUS STATISTICS

| Monolingual Corpus | No of Lines | No of Words |
|---|---|---|
| Sinhala [8] | 4,735,660 | 68,136,065 |
| Tamil [8] | 92,902 | 585,718 |
| English (Annual Reports/Gov Websites) | 23,118 | 205,307 |

#### B. List Integration

As per work by Carpuat and Diab [22] the best scores for SMT had been obtained by integrating the lists as a static training corpus. Therefore the same technique has been followed in the list integrations.

Bilingual Dictionaries and Glossaries can be easily found and these can introduce vocabulary to the TM.

It is a common observation that Named Entities (NE) will not exist in the training corpus. Therefore such terms would not be addressed by TM. As a result lists which contain, person names (E), addresses (D) have been considered to integrate as corpus. Specific to the official government documents domain, a curated list of designations and organizational names (F) have been used for training the TM.

TABLE VI. LIST STATISTICS

| Lexicon/Lists | List Ref | No of Terms | MT applicable |
|---|---|---|---|
| Dictionary (Si/Ta) | A | 19,132 | Si ↔ Ta |
| Dictionary (Si/En) | B | 23,722 | Si ↔ En |
| Trilingual Glossary | C | 24,261 | Si ↔ Ta ↔ En |
| Addresses List | D | 281,978 | Si ↔ Ta |
| Person Names List | E | 199,268 | Si ↔ Ta |
| Designations and Organization Names | F | 44,193 | Si ↔ Ta |

*C. Sinhala-English Data Augmentation*

Data sparseness or OOV is the main problem in MT as the TM would not be able to handle such words for translation in the context of SMT [7]. For morphologically rich languages that contain inflections this would be a greater challenge. Sinhala and Tamil languages are highly inflectional and hence the data sparseness would have a high impact in the MT process.

According to SMT research for Sinhala-Tamil [7] list/terminology integration had shown improved results. For local languages lexical resources such as dictionaries/glossaries can be found. The glossaries and dictionaries can be used in the TM to address the OOV problem to a certain extent. Therefore we integrated bilingual lists as static corpus [21], as a technique to address the OOV to improve the overall MT.

The lexicons and lists contain nouns in their base singular forms. Therefore the lists would translate the terms if they occur in the base form. If the same term appears in an inflected form, still the TM would not be able to translate, as outlined in Table I example.

To overcome this limitation, as a novel approach, the lists can be augmented and used in the TM. This is experimented with the Sinhala-English language pair.

Sinhala language is a morphologically rich language. As a highly inflected language, common nouns in Sinhala are inflected for number, definiteness and case. Sinhala nouns are also divided into animate and inanimate classes on the basis of their inflection. Animate nouns inflect for number (singular and plural), definiteness (definite and indefinite) and five cases (nominative, accusative, dative, genitive, and instrumental). Definiteness distinction applies only in the singular form of nouns [24].

Mainly for Sinhala animated common nouns the words are inflected based on five case markers as shown in Table VII. [15] It outlines how the grammatical case changes for animate common nouns.

TABLE VII. EXAMPLES FOR INFLECTIONS OF ANIMATE COMMON NOUNS

| Case | Singular | | | |
|---|---|---|---|---|
| | Masculine | | Feminine | |
| | Def | Indef | Def | Indef |
| Nominative | මිනිසා /min isa:/ | මිනිසෙක් /min isek/ | කෙල්ල /kella/ | කෙල්ලක් /kellak/ |
| Accusative | මිනිසා /min isa:/ | මිනිසෙකු /min iseku/ | කෙල්ල /kella/ | කෙල්ලක /kellaka/ |
| Dative | මිනිසාට /min isa:ṭə/ | මිනිසෙකුට /min isekuṭə/ | කෙල්ලට /kellaṭə/ | කෙල්ලකට /kellekuṭə/ |
| Genitive | මිනිසාගේ /min isa:ge:/ | මිනිසෙකුගේ /min isekuge:/ | කෙල්ලගේ /kellage/ | කෙල්ලකුගේ /kellakəge:/ |
| Instrumental | මිනිසාගෙන් /min isa:gen/ | මිනිසෙකුගෙන් /min isekugen/ | කෙල්ලගෙන් /kellakəgen/ | කෙල්ලකගෙන් /kellakəge:/ |

1. Nominative Case (ප්‍රථමා විභක්ති)

The dictionary term would be in the definite form. The indefinite term has been generated by inflecting the word with the respective suffix (ප්‍රත්‍ය).

The English nominative term has been generated by adding the articles 'the' and 'a/an' as preceding the term.

eg : මිනිසා (term) → **the man**, in Definitive form

මිනිසෙක් (inflected term with suffix එක්) → **a man**, in Indefinite form.

2. Accusative Case (කර්ම විභක්ති)

The Sinhala dictionary term would be in the definite form. For indefinite form term has been generated by inflecting the word with the respective suffix (ප්‍රත්‍ය) and in the objective (අනුක්ත) form.

The term for English accusative case rule would be adding the articles 'the' and 'a/an' as preceding the term.

eg : මිනිසා (term) → **the man**, in definitive form

මිනිසෙකු (suffix එකු) → **a man**, in Indefinite form.

3. Dative Case (සම්ප්‍රදාන විභක්ති)

The Sinhala dictionary term is modified by the suffix (ප්‍රත්‍ය) ie. 'ට' සහ 'උට' along with the definite and indefinite (එකු) suffixes with the word root.

The English dative term has been generated by adding the articles 'to the' and 'to a/an' as preceding the term.

eg : මිනිසාට (root+ආ+ට) → **to the man**, in definitive form

මිනිසෙකුට (root+ එක් +උට ) → **to a man**, in Indefinite form.

4. Genitive case (සම්බන්ධ විභක්ති)

The Sinhala dictionary term is modified by the suffix (ප්‍රත්‍ය ) ie. 'ගේ'. The definite and indefinite terms are generated by adding the character 'ගේ' as a suffix to its nominative case term.

The English Genitive term has been generated by adding the suffix ' 's ' to the nominative case term.

eg : මිනිසාගේ(root+ආ+ගේ) → **the man's**, in definitive form

මිනිසෙකුගේ (inflected term with suffix 'ගේ') → **a man's**, in Indefinite form.'

5. Instrumental Case (කරණ විභක්ති)

The Sinhala dictionary term is modified by the suffix ie. 'ගෙන්'. The definite and indefinite terms are generated by adding the 'ගෙන්' as a suffix to its nominative case term.

The English Instrumental case term has been generated by adding the suffix 'from the' or 'from a/an' to the English dictionary term.

eg : මිනිසාගෙන්(term+ගෙන්) → **from the man**,in definitive form

මිනිසෙකුගෙන් (inflected term with suffix 'ගෙන්') → **from a man**, in Indefinite form.

Based on the above rules the dictionary terms (nouns) and the glossary terms had been augmented and conducted the Si →En and En→Si SMT experiments.

Since the case-marker based morphological rules are valid for common nouns, from the dictionary the common singular nouns were considered. The common noun terms were 8,933 in the dictionary corpus. Based on these seed words the inflected terms were generated for both Sinhala and English sides.

For the glossary, the ending word of each glossary term was inflected based on case-markers. Finally the augmented parallel lists i.e. Sinhala–English were used in the SMT in Sinhala→English and English→Sinhala directions.

For Tamil language owing to limited resources, list augmentation has not been done.

Table VIII shows the dictionary and glossary original and augmented corpus statistics.

TABLE VIII. DICTIONARY AND GLOSSARY STATISTICS

| Lexicon | No of Terms |
|---|---|
| Dictionary (Si/En) | 23,722 |
| Augmented Dictionary (Si/En) | 45,767 |
| Glossary (Si/En) | 24,261 |
| Augmented Glossary (Si/En) | 83,433 |

*D. List with Filtration*

Although integration of dictionaries and glossaries is a low-cost technique for addressing OOV, it may not always work well. As the list contains term wise in the parallel corpus, [25] the TM would only know to translate such terms when occurs in the source sentence but the positioning of this term within the sentence would not be always correct. This has an impact on the translation output, as the word or phrase if not in the training corpus would stand on its own. The LM or the RM would not be able to decide the word position correctly.

Therefore the lists have been filtered based on the training corpus prior to training the model. Hence the re-ordering problem is expected to be minimised to the actual OOV which does not exist in the training corpus.

This technique has been applied for the Sinhala-English and Sinhala-Tamil SMT experiments.

IV. EXPERIMENTS AND RESULTS

The experimental setup was built using Moses [26] as a Phrase-Based Statistical Translation System (PBSMT) system.

As preprocessing tokenization, true casing (for English Language) and standard Moses cleaning scripts were applied to remove misaligned sentences ie. sentence pairs with a ratio of 1:9 to be removed from the parallel corpus. To generate the word alignment, Giza++ [27], was used with 'grow-diag-final-and' as the symmetrization heuristic and 'msd-bidirectional-fe' as the reordering technique. 'Good Turing' was used as the smoothing technique for the phrase table score smoothing.

In addition to phrase translation score; lexical translation scores, word and phrase penalties, and linear distortion were used as features in the TM, which are commonly used features [26]. An LM of order 3(3-gram) was created using SRILM [28].

The feature weights of each model were tuned using Minimum Error Rate Training (MERT) [29] on 100 best translations of 1000 sentences / phrases.

The baseline was set up with the training data of the parallel corpus for the TM and LM. The lexicons were integrated as corpus for the translation model.

The SMT experiments conducted for each language pair, according to the techniques described in section 3 are discussed in this section.

We use BLEU [30] as the MT evaluation metric . BLEU would be calculated based on the comparison of the n-grams of the translated text with the n-grams of the reference translation. To supplement this measure the out-of-vocabulary (OOV) rate is also stated to indicate the number of words that did not get translated by the respective SMT model.

### A. Sinhala-English SMT Experiments

For the Sinhala-English language pair the experiments were conducted as follows.

A1  Baseline
A2  Baseline+Dictionary
A3  Baseline+Augmented Dictionary
A4  Baseline+Glossary
A5  Baseline+Augmented Glossary
A6  Baseline+Augmented Dictionary+Glossary
A7  Baseline+ filtered Aug Dictionary filtered Glossary

Table IX summarizes the SMT experiments conducted for the Sinhala-English pair along with the respective BLEU scores and OOV rates.

TABLE IX. Sinhala-English SMT Experimental Results.

| Ref. | Si → En | | En → Si | |
|---|---|---|---|---|
|  | BLEU | OOV | BLEU | OOV |
| A1 | 20.12 | 1382 | 17.08 | 1074 |
| A2 | 20.89 | 1115 | 17.25 | 838 |
| A3 | 21.15 | 1078 | 17.72 | 798 |
| A4 | 21.02 | 1112 | 17.76 | 780 |
| A5 | 20.59 | 1135 | 17.32 | 835 |
| A6 | 21.18 | 1065 | 17.89 | 787 |
| A7 | **21.20** | **1003** | **18.03** | **770** |

With the Sinhala-English SMT experiments focus has been given to list integration with augmentation. It is observed that the augmented dictionary (A3) integration increases the BLEU score in both directions. However the augmented glossary integration (A5) does not provide an increase in BLEU score for both directions. Nevertheless the OOV has reduced in both experiments. The glossary terms consist of multiple words. The observation is that the augmentation at word level is more reliable than augmentation at the phrase level.

However the list filtration increases the BLEU score in all experiments reducing the OOV as well.

### B. Sinhala-Tamil SMT Experiments

For the Sinhala-Tamil language pair the experiments that have been conducted are as follows.

B1 Baseline

B2 Baseline+Dictionary

B3 Baseline+Glossary

B4 Baseline+Dictionary+Glossary+ListD+ListE+ListF

B5 Baseline+FiltDic-FiltGl-FiltListD-FiltListE-FiltListF

Table X summarizes the SMT experimental results conducted for the Sinhala-Tamil pair and the respective BLEU scores and the OOV count.

TABLE X. Sinhala-Tamil SMT Experimental Results.

| Ref. | Si → Ta | | Ta → Si | |
|---|---|---|---|---|
|  | BLEU | OOV | BLEU | OOV |
| B1 | 22.46 | 965 | 26.88 | 1993 |
| B2 | 22.78 | 870 | 27.02 | 1910 |
| B3 | 22.77 | 871 | 27.00 | 1881 |
| B4 | 23.36 | 704 | 27.67 | 1711 |
| B5 | **23.40** | **692** | **27.79** | **1866** |

In the Sinhala–Tamil experiments the main focus was to analyse the improvement in the translation output with list integration and filtration. Overall in the Sinhala→Tamil direction there's an improvement of **0.94** and in the Tamil→Sinhala direction an improvement of **0.91.** In line with the BLEU scores the OOV has also reduced. Further the list integration with filtration has given rise to an increase in BLEU score which further justifies that filtered list integration performs better than by integrating the full list for training.

## C. Tamil-English SMT Experiments

Table XI summarizes the SMT experiments conducted for the Tamil-English language pairs along with the respective BLEU scores.

TABLE XI. ENGLISH-TAMIL SMT EXPERIMENTAL RESULTS.

| Ref. | Experiments | En → Ta | | Ta → En | |
|---|---|---|---|---|---|
| | | BLEU | OOV | BLEU | OOV |
| C1 | Baseline | 16.70 | 3020 | 12.98 | 1532 |
| C2 | Baseline + Glossary | **17.13** | 2690 | **13.04** | 1496 |

For Tamil–English translation it is evident that integration of the glossary as a list enhances the SMT score by **0.33-0.06** in the respective directions.

## V. CONCLUSION AND FUTURE WORK

This paper has focused on exploiting bilingual list integrations towards addressing the OOV problem in SMT for Sinhala-Tamil-English languages.

Extending from the technique of integrating the full lists as they are, we have observed that augmenting the list based on case-markers gives better performance for the translation output. In Sinhala-English MT, it was observed that the dictionary list augmentation in both directions directly improves the BLEU and reduces the OOV. However the augmented glossary although did not increase the BLEU, shows it has reduced the OOV which still results in an qualitative improvement on the translation output.

Based on the observation made in the SMT for Sinhala-English TMs, we believe by including the augmented terms in a synthetic sentence could improve the BLEU score as an alternative to integrating the augmented term as a list into the TM. The work on synthetic data generation [13] can be extended to incorporate the augmented list terms.
Further we believe the same augmentation technique can be applied for the Tamil language, given that suitable resources are found.

## VI. ACKNOWLEDGEMENTS

This research was supported by the Accelerating Higher Education Expansion and Development (AHEAD) Operation of the Ministry of Education funded by the World Bank.